\documentclass[conference]{IEEEtran}

\usepackage{cite}
\usepackage{amsmath,amssymb,amsfonts}
\usepackage{algorithmic}
\usepackage{graphicx}
\usepackage{textcomp}
\usepackage{xcolor}
\usepackage{subcaption}
\usepackage{bm}
\usepackage{booktabs} 
\usepackage{tabu}
\usepackage{cellspace} 

\usepackage{amsmath}
\usepackage{amsfonts}
\usepackage{amssymb}
\usepackage{hyperref}

\def\BibTeX{{\rm B\kern-.05em{\sc i\kern-.025em b}\kern-.08em
    T\kern-.1667em\lower.7ex\hbox{E}\kern-.125emX}}
\begin{document}

\title{A3: Active Adversarial Alignment for Source-Free Domain Adaptation}

\author{
\IEEEauthorblockN{Chrisantus Eze and Christopher Crick}
\IEEEauthorblockA{Department of Computer Science,
Oklahoma State University, Stillwater, USA \\
chrisantus.eze@okstate.edu, chriscrick@okstate.edu
}
\thanks{Corresponding author: Chrisantus Eze (chrisantus.eze@okstate.edu)}
}


\maketitle

\begin{abstract}
Unsupervised domain adaptation (UDA) aims to transfer knowledge from a
labeled source domain to an unlabeled target domain. Recent works have
focused on source-free UDA, where only target data is available. This
is challenging as models rely on noisy pseudo-labels and struggle with distribution shifts. We propose Active Adversarial Alignment (A3), a novel framework combining self-supervised learning, adversarial
training, and active learning for robust source-free UDA. A3 actively
samples informative and diverse data using an acquisition function for
training. It adapts models via adversarial losses and consistency
regularization, aligning distributions without source data access. A3
advances source-free UDA through its synergistic integration of active
and adversarial learning for effective domain alignment and noise
reduction. Our approach significantly advances
state-of-the-art methods, achieving 4.1\% on Office-31, 11.7\% on Office-Home, and 10.6\% on DomainNet accuracy improvements. Source code: \url{https://github.com/chrisantuseze/active-self-pretraining}
\end{abstract}

\begin{IEEEkeywords}
domain adaptation, self-supervised learning, adversarial learning, source-free adaptation.
\end{IEEEkeywords}

\section{Introduction}\label{sec:intro}

\begin{figure*}[htbp] 
\centering
\includegraphics[scale=0.34]{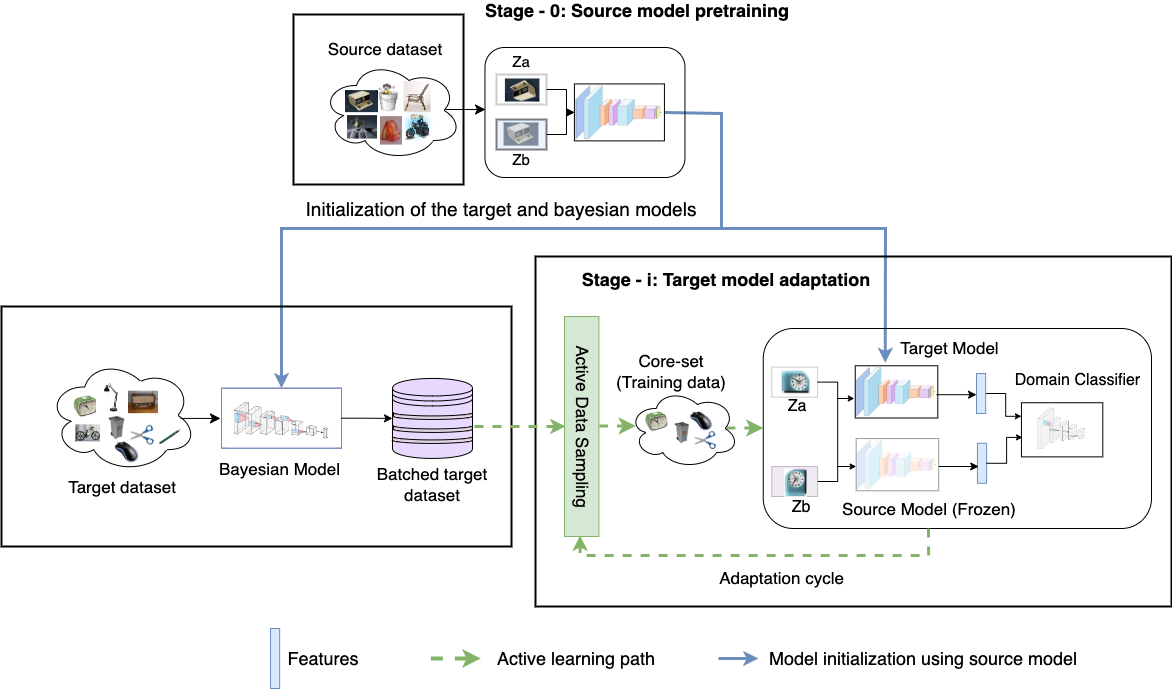} 
\caption{Our framework involves two main phases: the source model pretraining and the model adaptation. We begin by adopting a training regime for the source model, which is initialized with a pre-trained ImageNet model \cite{deng2009imagenet}. This is referred to as stage-0 in the multi-stage adaptation process. The second phase encompasses n-1 stages of active learning. During this phase, a core-set is constructed and the Bayesian model uncertainty is estimated. At each iteration, the core-set is updated with the top-k informative samples from the training pool. These samples are then used to retrain the Bayesian and target models until the data sampling budget is exhausted. Refer to Section \ref{sec:proposed_method} for
  further details on target domain alignment. Za and Zb denote the two
  augmentations of the input used for self-supervised training of the
  source and target models.}
\label{fig:architecture}
\end{figure*}

Unsupervised domain adaptation (UDA) addresses the poor model
performance that arises due to domain shift \cite{pan2010survey} by
leveraging labeled data from a source domain to train models that
generalize to an unlabeled target domain. However, standard UDA
techniques require access to source data, which might not be feasible
due to privacy or computational resource concerns. This paper tackles
the challenging problem of source-free unsupervised domain adaptation
(SFUDA), where only target data is available without its label. Recent SFUDA methods \cite{liang2020we,hou2021visualizing,ishii2021source,jing2022variational},
assume access to a pre-trained source model.

Various approaches have been proposed to address domain shifts in
source-to-target domain adaptation. In a semi-supervised setting, works like
\cite{french2017self,shu2018dirt,zou2019confidence,yan2021augmented}
address the problem using model regularization techniques and
self-training with pseudo-labels. Another line of work focuses on
aligning source and target feature distributions, with notable works
including
\cite{su2020active,long2018conditional,ganin2015unsupervised}. These
approaches design adversarial domain discriminators in parallel with
the classification head.

The idea of designing pseudo-labels for training the target model has
been prevalent in recent literature. SHOT \cite{liang2020we} refines
pseudo-labels with a prototype classifier and fine-tunes the feature
extractor with a model regularization term maximizing mutual
information. The work done in \cite{yan2021augmented} introduces an
augmented self-labeling scheme to improve pseudo-labels and retrain
the target model.

Despite the benefits of self-labeling schemes, they face challenges
such as noisy pseudo-labels since they rely on predictions made by a
model trained on the source domain to label target domain samples. In
addition, due to the problem of prior initialization
\cite{Houlsby2011BayesianAL,yan2021augmented}, there might be limited
data used to initialize the target for pseudo-labeling. To address
these issues, we propose Active Adversarial Alignment for Source-Free Domain Adaptation (A3), a novel
approach for source-free unsupervised domain adaptation as shown in
Fig. \ref{fig:architecture}. A key contribution of A3 is an active
learning strategy that uses an acquisition function to carefully
select the most informative and diverse target samples to build a
core-set for training the target model. Using an acquisition
function based solely on uncertainty or diversity sampling tends to be
less effective for active domain adaptation
\cite{prabhu2021active,su2020active}. Therefore, we adopted a hybrid
acquisition strategy that combines uncertainty and diversity sampling
to identify both informative and representative samples from dense
regions of the feature space.

Furthermore, we adapt the source model to the target domain using
adversarial and consistency losses that encourage learning
domain-invariant features without source data. Specifically, we employ
a domain adversarial loss which trains a domain classifier to
distinguish between target embeddings generated from the source and target models. By using a gradient
reversal layer, we ultimately confuse this classifier thereby reducing
domain divergence. Additionally, we incorporate a virtual adversarial
loss \cite{miyato2018virtual} which locally perturbs embeddings to
maximize prediction change and enforce local Lipschitz smoothness. The
virtual adversarial loss acts as a regularization technique to prevent
overfitting and encourage robustness. Together, the domain adversarial
and virtual adversarial losses perform global and local distribution
alignment to facilitate effective adaptation. Further, we utilize a
swap prediction loss for self-supervision and an entropy minimization
term to prevent target overfitting. We summarize our contributions as follows:
\begin{enumerate}
    \item We propose a new framework for domain alignment by jointly
      training the target model and a domain classifier to enable the
      target model to produce domain-invariant features compelled by
      adversarial and regularization losses.
    \item To address the problem of noisy pseudo-labels in supervised
      and semi-supervised domain adaptation, we introduce A3, an
      active self-supervised adversarial training strategy to achieve
      source-free and target-label-free domain alignment.
    \item To the best of our knowledge, this is the first
      comprehensive work that combines self-supervised learning,
      adversarial training, and active learning to achieve source-free
      unsupervised domain adaptation. We also performed extensive
      evaluations on benchmark datasets achieving impressive
      state-of-the-art performance.
\end{enumerate}

\begin{figure*}
    \centering
    \includegraphics[scale=0.43]{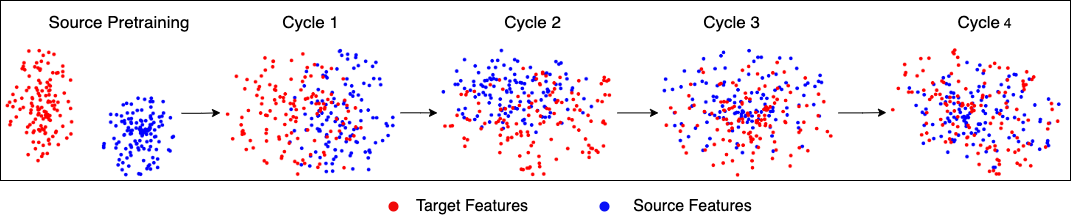}
    \caption{T-SNE plot of the learned source features and the target features at each active learning adaptation cycle.}
\label{fig:visualization}
\end{figure*}

\section{Related Work} \label{sec:relatedwork}

Settles' comprehensive survey \cite{settles2009active} delves into
active learning, exploring acquisition functions such as information
theoretical approaches \cite{6795562} and uncertainty-based methods
\cite{tong2001support}. The CLUE framework
\cite{prabhu2021active} introduces uncertainty-weighted clustering
for diverse instance selection under domain shifts. The synergy of
BALD \cite{Houlsby2011BayesianAL} with deep neural networks, amplifies acquisition
performance. Batch sampling strategies, such as those involving BALD
\cite{kirsch2019batchbald} and core-set approximations
\cite{Pinsler2019BayesianBA}, address efficiency concerns posed by
computational challenges.

Additionally, Active Domain Adaptation (ADA) optimizes domain
adaptation by strategically selecting samples. \cite{rai2010domain}
introduced ADA, later adapted to image classification as AADA
\cite{su2020active}. TQS \cite{fu2021transferable} and CLUE
\cite{prabhu2021active} emphasize uncertainty-based sample selection,
while S3VAADA \cite{rangwani2021s3vaada} incorporates vulnerability,
diversity, and representativeness.  Inspired by BALD and CLUE, we adopted a hybrid acquisition function that jointly captures both uncertainty and diversity of the samples.

The Gradual Source Domain Expansion (GSDE) approach \cite{westfechtel2024gradual} presents a method for mitigating early alignment errors in Unsupervised Domain Adaptation (UDA) by progressively integrating pseudo-source data from high-scoring target samples. This method emphasizes the incremental alignment of source and target domains over multiple training iterations. In contrast to this gradual expansion method, our approach utilizes active learning to sample the most diverse and representative instances upfront, alongside adversarial learning and model regularization, to better ensure domain invariance from the outset.

In another related work, Local Context-Aware Active Domain Adaptation (LADA) \cite{sun2023local}, the authors proposed an active selection criterion based on local inconsistency in model predictions, focusing on uncertain regions. LADA uses a Progressive Anchor-set Augmentation (PAA) module to handle the small size of queried data, supplementing labeled target data with pseudo-labeled confident neighbors. Our method differs by emphasizing not just uncertainty but also diversity in sample selection, and by incorporating adversarial learning to improve domain adaptation, coupled with model regularization to avoid overfitting.

Over time, there has been growing interest in aligning domains through
adversarial training. Various approaches have been proposed:
\cite{tzeng2014deep} introduced an adversarial loss alongside a
classification loss for learning domain-invariant features. Similarly,
\cite{tzeng2017adversarial} used an adversarial loss to train a
target model to deceive a discriminator in stages. Adversarial
training, as noted by \cite{saito2019semi}, helps prevent overfitting
of the target classifier to unlabeled target data by maximizing
classifier uncertainty. Our work leverages a domain adversary to
discriminate between embeddings from two domains, ensuring the target
model learns to produce invariant features.

\section{Proposed Method}
\label{sec:proposed_method}
In this section, we introduce our novel approach, specifically designed for source-free
unsupervised domain adaptation. Our primary objectives include reducing annotation costs through
the utilization of self-supervised learning and active learning for
iterative model adaptation. To address these goals, we introduce and
integrate various domain adaptation techniques, including a swap
prediction loss, along with the introduced domain adversarial loss,
virtual adversarial loss, and entropy minimization loss. In addition,
inspired by BALD \cite{Houlsby2011BayesianAL} and CLUE
\cite{prabhu2021active}, we adopted a hybrid acquisition function
that jointly captures both uncertainty and diversity of the
samples.

In the first part of this section, we present the self-supervised learning technique we adopted for pretraining the source model. In the second part, we discuss our acquisition
strategy, which involves iteratively sampling low-entropy and diverse
instances from the target pool to train the target model. Finally, in the
third part, we introduce adversarial losses and model regularization
techniques, providing a detailed overview of our methodology. This
structured presentation ensures a clear understanding of the proposed
A3 method and its components.

The combined benefit of this iterative process, adversarial losses, and model regularization enhances the model's
performance in adapting to the target domain, ensuring that the target
model learns to produce invariant features. The T-SNE \cite{van2008visualizing} plot for these features can be visualized Fig. \ref{fig:visualization}.

\subsection{Self-Supervised Pretraining}
We adopt a self-supervised learning scheme inspired by SwAV
\cite{caron2020unsupervised}, aiming to maximize the similarity
between positive pairs and minimize the similarity between negative
pairs. This involves contrasting multiple image views by comparing
their cluster assignments instead of their features as used in
\cite{chen2020simple}. The method groups data
into clusters while maintaining consistent cluster assignments across
different augmentations or "views" of the same image. Initially,
"codes" are generated by associating features with prototype
vectors. Subsequently, a "swap" prediction task is solved, where
codes obtained from one augmentation are predicted using the
other. This ensures that the model doesn't directly compare image
features. For two image features, \(z_t\) and \(z_s\), obtained from
distinct augmentations of the same image, their codes \(q_t\) and
\(q_s\) are derived by matching these features to a set of \(K\)
prototypes \(\{c_1, \ldots, c_K\}\). The swap prediction loss function
is given by

\begin{equation}
    L(z_t, z_s) = \ell(z_t, q_s) + \ell(z_s, q_t)
\end{equation}

where the function $\ell(z, q)$ measures the consistency between
features $z$ and a code $q$. This is expanded into

\begin{equation}
    \begin{split}
         \mathcal{L}_{swap} = -\frac{1}{N} \sum_{n=1}^{N} \sum_{s,t \sim T} [ \frac{1}{\tau} z_{nt}^TCq_{ns} + \frac{1}{\tau} z_{ns}^TCq_{nt} \\
         - \log \sum_{k=1}^{K} \exp\ \left( \frac{z_{nt}^Tc_k}{\tau} \right) - \log \sum_{k=1}^{K} \exp\ \left( \frac{z_{ns}^Tc_k}{\tau} \right) ]
    \end{split}
    \label{eq:swav_loss}
\end{equation}

where $z_{nt}$ and $z_{ns}$ are features from the two compared image augmentations,
and $q_{nt}$ and $q_{ns}$ are their intermediate codes, with $Cq_{ns}$
and $Cq_{nt}$ representing their prototypes. This self-supervised
learning approach is used for training both the source and target
models.

\subsection{Active Data Sampling}
\label{sec:active_sampling}
In the context of active learning, determining the uncertainty or
informativeness of a sample is crucial for selecting the next sample
to query. This is achieved through an acquisition function employed by
the active learning (AL) system. Various works in the literature have
proposed different acquisition functions, as extensively discussed in
\cite{Gal2016UncertaintyID}.

In the active learning process, given an unlabeled dataset
$\mathcal{X}_p$ and the current training pool $\mathcal{D}_o$ serving
as the core-set, a bayesian model $\mathcal{M}$ with parameters
$\omega \sim p(\omega|\mathcal{D}_o)$ as inputs, the acquisition
function ranks batch samples based on model uncertainty and sample
representativeness. The system then selects highly informative and
representative samples from the batch
\cite{kirsch2019batchbald}. 

\subsubsection{Acquisition Strategy}
BALD \cite{Houlsby2011BayesianAL}, an uncertainty sampling strategy,
determines the optimal unlabeled sample, denoted as $x^*$, by
evaluating the mutual information between predictions and the model
posterior. While BALD primarily focuses on exploitative uncertainty
sampling, a desire for exploration of diverse instances can be
incorporated by introducing a distance-based diversity reward, just like in CLUE \cite{prabhu2021active}. This extends the acquisition
function to consider under-explored regions in the input space through
cluster-based distances.

To implement this strategy, the indices of the weighted instances are
determined, and the Euclidean distances of these instances are sorted
in descending order. The goal is to sample the top-k instances with
the least uncertainty and high diversity based on their Euclidean
distance for training the target model. For convenience, we will refer
to this measure of uncertainty and diversity as the A3 score.

\subsubsection{Core-set Construction}
\label{sec:core_set_constr}
We aim to overcome a limitation of BALD arising from the use of an
uninformative prior, which results from poor initialization of the
core-set. To address this, we initialize a model using the source
model to solve a pretext classification task where the labels are
given by $y \in \{0, 90, 180, 270\}$ which represents possible
rotation angles in degrees for the augmentations applied to the
sample following the work done in \cite{yi2022using}. This pretext-task model serves as the Bayesian model,
$\mathcal{M}$ used to construct a data pool $\mathcal{D}_o$. During
this process, we perform inference over the parameters $\theta$ to
obtain the posterior distribution $p(\theta|\mathcal{D}_o)$
\cite{Pinsler2019BayesianBA}, sampled by the A3 score. The data pool
is then sorted by the A3 score of the samples and grouped into $n$
batches.

This approach not only facilitates optimal sample querying but also
addresses the cold-start problem inherent in active learning
\cite{yi2022using}. Our acquisition strategy given in the previous
section is utilized to select the top-k samples from the first batch
of the data pool, forming the core-set for training the target
model. Subsequently, $\mathcal{M}$ is retrained on this core-set, and
the iterative process continues. The next batch of the data pool is
equally passed through the acquisition function and the top-k samples
are added to the core-set. This iterative cycle involves retraining
the target model and $\mathcal{M}$, sampling instances from the next
batch using the acquisition function, expanding the core-set, and so
on until the sampling budget is exhausted.

This iterative training and data sampling technique is visually
depicted in Fig. \ref{fig:architecture}.

\subsection{Domain Alignment}
\label{sec:domain_alignment}
The objective of domain adaptation is to train a model that is
invariant across domains, capable of delivering accurate predictions
in both the source and target domains. Drawing upon the representative
and informative data selected through our active sampling technique
detailed in Section \ref{sec:active_sampling}, we introduce a novel
training routine for self-supervised based SFUDA. This routine
incorporates two adversarial losses—domain adversarial loss and
virtual adversarial loss—to drive the target model toward generating
invariant representations. Additionally, we adopt a swap prediction
loss (\ref{eq:swav_loss}) inspired by SwAV
\cite{caron2020unsupervised} and an entropy minimization loss
\cite{grandvalet2004semi} as regularization techniques. These
measures collectively aim to mitigate the overfitting of representations
to the target domain and address the divergence between predictions in
the source and target domains.

\subsubsection{Model Regularization}
\label{sec:model_reg}
Incorporating the swap prediction loss outlined in \ref{eq:swav_loss} as a regularization technique in our approach, we
guide the model to learn meaningful representations by predicting
relationships between augmented instances. Beyond the swap prediction
loss, we introduce an additional regularization term known as the
conditional entropy minimization loss \cite{yan2021augmented,
  grandvalet2004semi}. This term constrains the model,
preventing overfitting to the target domain and mitigating the
emergence of spurious correlations
\cite{grandvalet2004semi}. This is formally expressed as

\begin{equation}
    \mathcal{L}_\texttt{ent} = -\frac{1}{N} \sum_{i=1}^{N} \sum_{k=1}^{K} f_k(x_i; \theta) \log f_k(x_i; \theta)
    \label{eq:entropy_min}
\end{equation}

where $f(x; \theta)$ represents the output of the model parameterized
by $\theta$. This regularization term is introduced to address the
expectation that optimal decision boundaries should be distanced from
the data-dense regions of the samples, as emphasized in
\cite{shu2018dirt}. This aligns with the clustering assumptions,
asserting that target samples form clusters, and the samples within
the same cluster belong to the same class.

\subsubsection{Adversarial Losses}
\label{sec:adversarial_losses}
Moreover, both \cite{grandvalet2004semi} and \cite{shu2018dirt}
observed that the assumption in \ref{eq:entropy_min} holds
true only if the model is locally Lipschitz. To ensure this, we
incorporate the modified virtual adversarial loss (VAT) tailored for a
self-supervised learning setting, akin to the approach in
\cite{yan2021augmented}. This VAT loss minimizes the divergence
between predictions on clean samples vs those with small
perturbations. This smooths the decision boundary and improves
robustness:

\begin{equation}
    \mathcal{L}_\texttt{vat} = \mathcal{D}[f(x), f(x + r_{vadv})]
    \label{eq:vat}
\end{equation}

where $f(x)$ is the model output embedding for input $x$ and
$r_{vadv}$ is the computed VAT perturbation to maximize divergence
between $f(x)$ and $f(x + r_{vadv})$. The $\mathcal{D}$ is the KL
divergence. We therefore aim to minimize $\mathcal{L}_{vat}$ to
enforce local smoothness of the model output \cite{shu2018dirt} and
also aid the model generalization to the target domain while still
retaining knowledge from the source domain without catastrophic
forgetting \cite{kirkpatrick2017overcoming}.

Additionally, we introduce a second adversarial loss to learn an embedding space where the domain adversary
cannot reliably predict the domain from the embeddings. In this
context, a domain classifier $\mathcal{D}(f(x))$ is co-trained with
the target model, using target embeddings extracted separately from
the source and target models to predict the model they originated
from. To optimize the domain classifier, we employ a gradient reversal
layer \cite{ganin2015unsupervised}, which flips the sign of the
gradients during backpropagation. This adversarial learning approach
makes the features challenging for domain prediction, and encourages
the target model to learn to produce domain-invariant features. The
domain adversarial loss, DAL, is expressed as

\begin{equation}
    \begin{split}
        \mathcal{L}_\texttt{dal} = \mathbb{E}_{f(x_t) \sim F_s} [-\log \mathcal{D}(f_s(x_t))] \\
        + \ \mathbb{E}_{f(x_t) \sim F_t} [-\log(1 - \mathcal{D}(f_t(x_t)))]
    \end{split}
    \label{eq:dal}
\end{equation}

where $\mathcal{D}$ is the domain classifier, $x_t$ is the target
samples and $F_s$ and $F_t$ are the distributions of extracted source
and target embeddings respectively.

The key distinctions between Domain Adversarial Loss (DAL), as
described in \ref{eq:dal}, and Virtual Adversarial Loss
(VAT), as described in \ref{eq:vat}, lie in their underlying
approaches. DAL focuses on training a domain classifier to
differentiate embeddings from the source and target models, leveraging
the distinction between the two domains. On the other hand, VAT
generates perturbations around an input to maximize the prediction
change. DAL primarily aims for global alignment of source and target
distributions, whereas VAT regularization ensures that individual
sample predictions remain locally invariant. Additionally, DAL is
designed to reduce the H-divergence between domains, providing an
upper bound on the target error. In contrast, VAT ensures the local
Lipschitz constraint necessary for reliable empirical estimation.

The overall loss for the model is expressed as

\begin{equation}
    \mathcal{L}  = \mathcal{L}_\texttt{swap} + \lambda_1\mathcal{L}_\texttt{dal} + \lambda_2(\mathcal{L}_\texttt{ent} + \mathcal{L}_\texttt{vat})
\end{equation}

In the overall loss for the model, $\lambda_1$ represents a
regularization hyperparameter for the domain adversarial loss, while
$\lambda_2$ serves as trade-off hyperparameter shared by the entropy
loss and the virtual adversarial loss, as suggested by
\cite{shu2018dirt} and \cite{li2020model}.

The effectiveness of A3 can be attributed to its multi-faceted approach to domain alignment. The active learning component ensures that the most informative target samples are utilized, reducing the impact of noisy or irrelevant data. Meanwhile, the adversarial training encourages the model to learn domain-invariant features, bridging the gap between source and target distributions. The self-supervised learning aspect further enhances the model's ability to capture meaningful representations without relying on target labels.
This combination is particularly effective for SFUDA tasks with a reasonable degree of shared structure between domains, even if the surface-level statistics differ. For instance, in image classification tasks across different photo styles (e.g., Amazon product images to real-world images), the underlying object structures remain consistent, allowing A3 to leverage these commonalities effectively.

\section{Experiments}
\label{sec:experiments}
In this section, we conduct rigorous evaluations of our approach to
investigate and prove A3's robustness and effectiveness in carrying
out source-free domain adaptation.

\subsection{Datasets}
\label{sec:datasets}
Following the baselines, we evaluated A3 on various benchmark datasets
that represent different visual domains to gauge its robustness and
generalizability.

The \textbf{Office-31} \cite{saenko2010adapting} dataset has 4700
images in 31 categories from Amazon (A), DSLR (D), and Webcam (W)
domains, while \textbf{Office-Home} \cite{venkateswara2017deep} has
15500 images in 65 categories from Artistic (A), Clip-Art (C), Product
(P), and Real-World (R) domains.  Additionally, we evaluated A3 on the
challenging \textbf{DomainNet} \cite{peng2019moment} dataset which
contains images from six domains with 345 categories each. However,
following the baselines, our evaluations of A3 were focused on four out
of the six domains: \textit{sketch}, \textit{clipart},
\textit{painting}, and \textit{real} which shows the model's
generalization between synthetic and real domains. 

\textbf{Implementation Details:} We adhered to established practices
by selecting ResNet-50 \cite{he2016deep} as the architecture for our
target model, pretrained on ImageNet \cite{deng2009imagenet}. The
network configuration closely mirrored that of SwAV
\cite{caron2020unsupervised}, with some custom adjustments. The
domain discriminator consisted of two layers and a classification head
with a single neuron for binary classification. Our Bayesian model,
integral to the active sampling process, utilized a ResNet-50 backbone
with a classification head tailored for 4-class classification,
corresponding to the four distinct input augmentations. For the
Bayesian model, Stochastic Gradient Descent (SGD) with a learning rate
of 0.1 and a multi-step learning rate scheduler was employed. We
conducted four active learning cycles, allocating equal sampling
budgets at each stage. In the self-supervised pretraining phase, SGD
with a learning rate of 1e-4 and a cosine learning rate scheduler were
utilized. Both self-supervised pretraining and the Bayesian model
implementation incorporated a momentum of 0.9 and a weight decay of
1e-6.

\subsection{Evaluations}
\label{sec:evaluations}
We compare our proposed framework, A3 with baselines on the benchmark
datasets highlighted in Section \ref{sec:datasets}. To showcase the
efficacy of A3, we compare it to the following baselines: ResNet-50
\cite{he2016deep}, SHOT \cite{liang2020we}, UAN
\cite{you2019universal}, InstaPBM \cite{li2020rethinking}, Sentry
\cite{prabhu2021sentry}, FixBi \cite{na2021fixbi}, GSDE
\cite{westfechtel2024gradual}, and LAS (LADA) \cite{sun2023local}

As shown in Table \ref{tab:office_31}, our A3 outperforms
existing state-of-the-art techniques on all the adaptation tasks on the Office-31 dataset with an improvement of 4.1\%. For the Office-Home dataset on the other, Table \ref{tab:office_home} shows that A3 demonstrates highly
competitive performance on 10 out of 12 transfer tasks, achieving
significant improvements in all tasks excluding \textbf{A$\to$C} and \textbf{R$\to$P}. Again, we outperform LAS \cite{sun2023local}, the next best-performing method with an average increase of 11.7\% accuracy. Finally, we show that A3 just as in the previous baseline datasets outperforms existing methods on DomainNet with an average improvement of 10.6\% accuracy. This is shown in Table \ref{tab:domainnet}.

While A3 shows overall improvements, its performance varies across different transfer tasks. For instance, on Office-31, A3 excels in the \textbf{D$\to$A} task (94.8\% accuracy), likely due to the shared low-level features between DSLR and Amazon domains. However, it shows more modest gains on the \textbf{A$\to$W} task (98.5\%), possibly due to the larger domain gap between Amazon and Webcam images.
On Office-Home, A3 demonstrates particular strength in transfers involving the Art domain (e.g. \textbf{A$\to$R}: 99.6\%, \textbf{R$\to$A}: 98.9\%). This suggests that our method effectively bridges the gap between realistic and artistic representations. However, the improvement is less pronounced or not evident for some intra-realistic transfers (e.g. \textbf{R$\to$P}: 93.7\%), indicating room for further optimization in scenarios with subtle domain shifts.

\begin{table}[htbp]
\small
  \centering
    \caption{An ablation study using various A3 variants on the
      A$\to$W task. Each method utilized the pretrained target model
      for classification on both the source and target
      datasets.}\label{tab:ablation}

  \begin{tabular}{lll}
   \toprule
    \bfseries Method & \bfseries Source (A) & \bfseries Target (W)  \\
    \midrule
     Hybrid & 98.3 & 98.5 \\
     Uncertainty Only & 98.2 & 98.3 \\
     Random & 94.2 & 95.9 \\
     \hline
     Consolidated & 98.3 & 98.5 \\
     DAL + VAT Only & 97.9 & 98.4 \\
     Entropy Only & 96.6 & 98.1 \\
    \bottomrule
  \end{tabular}
\end{table}

\begin{table*}[htbp]
\small
  \centering
    \caption{Accuracy (\%) on Office-31 for unsupervised domain
      adaptation (ResNet-50). The best accuracy is indicated in bold,
      while second best is underlined.}\label{tab:office_31}

  \begin{tabular}{llllllll}
   \toprule
    \bfseries Method & \bfseries A$\to$D & \bfseries A$\to$W & \bfseries D$\to$A & \bfseries D$\to$W & \bfseries W$\to$A & \bfseries W$\to$D & \bfseries Avg \\
    \midrule
     ResNet-50 & 68.9 & 68.4 & 62.5 & 96.7 & 60.7 & 99.3 & 76.1 \\
     \hline
     CAN & 95.0 & 94.5 & 78.0 & 99.1 & 77.0 & 99.8 & 90.6 \\
     SHOT & 94.0 & 90.1 & 74.7 & 98.4 & 74.3 & 99.9 & 88.6 \\
     FixBi & 95.0 & 96.1 & 78.7 & 99.3 & 79.4 & \textbf{100.} & 91.4 \\
     GSDE & 96.7 & 96.9 & 78.3 & 98.8 & 79.2 & \textbf{100.} & 91.7 \\
     LAS & \underline{96.9} & \underline{97.6} & \underline{84.2} & \textbf{100.} & \underline{86.0} & \textbf{100.} & \underline{94.1} \\
     \hline
     A3 (Ours) & \textbf{100.} & \textbf{98.5} & \textbf{94.8} & \textbf{100.} & \textbf{94.5} & \textbf{100.} & \textbf{98.0} \\
    \bottomrule
  \end{tabular}
\end{table*}

\begin{table*}[htbp]
\small
  \centering
  \caption{ Accuracy (\%) on Office-Home for unsupervised domain
    adaptation (ResNet-50). The best accuracy is indicated in bold,
    while the second best is underlined.}\label{tab:office_home}
  \begin{tabular}{lllllllllllllll}
   \toprule
    \bfseries Method & \bfseries A$\to$C & \bfseries A$\to$P & \bfseries A$\to$R & \bfseries C$\to$A & \bfseries C$\to$P & \bfseries C$\to$R & \bfseries P$\to$A & \bfseries P$\to$C & \bfseries P$\to$R & \bfseries R$\to$A & \bfseries R$\to$C & \bfseries R$\to$P & \bfseries Avg \\
    \midrule
    ResNet-50 & 34.9 & 50.0 & 58.0 & 37.4 & 41.9 & 46.2 & 38.5 & 31.2 & 60.4 & 53.9 & 41.2 & 59.9 & 46.1 \\
    \hline
     FixBi & 58.1 & 77.3 & 80.4 & 67.7 & 79.5 & 78.1 & 65.8 & 57.9 & 81.7 & 76.4 & 62.9 & 86.7 & 72.7 \\
     DCAN & 54.5 & 75.7 & 81.2 & 67.4 & 74.0 & 76.3 & 67.4 & 52.7 & 80.6 & 74.1 & 59.1 & 83.5 & 70.5 \\
     SHOT & 57.1 & 78.1 & 81.5 & 68.0 & 78.2 & 78.1 & 67.4 & 54.9 & 82.2 & 73.3 & 58.8 & 84.3 & 71.8 \\
     Sentry & 61.8 & 77.4 & 80.1 & 66.3 & 71.6 & 74.7 & 66.8 & 63.0 & 80.9 & 74.0 & 66.3 & 84.1 & 72.2 \\
     GSDE & 57.8 & 80.2 & 81.9 & 71.3 & 78.9 & 80.5 & 67.4 & 57.2 & 84.0 & 76.1 & 62.5 & 85.7 & 73.6 \\
     LAS & \textbf{77.8} & \underline{91.8} & \underline{88.4} & \underline{77.7} & \underline{91.5} & \underline{87.7} & \underline{78.1} & \underline{79.1} & \underline{89.5} & \underline{83.4} & \underline{79.8} & \textbf{94.1} & \underline{84.9} \\
     \hline
     A3 (Ours) & \underline{77.7} & \textbf{99.7} & \textbf{99.6} & \textbf{99.6} & \textbf{99.3} & \textbf{96.6} & \textbf{79.1} & \textbf{97.9} & \textbf{96.8} & \textbf{98.9} & \textbf{97.9} & \underline{93.7} & \textbf{94.8} \\
    \bottomrule
  \end{tabular}
\end{table*}

\begin{table*}[htbp]
\small
  \centering
  \caption{ Accuracy (\%) on DomainNet for unsupervised domain
    adaptation (ResNet-50). The best accuracy is indicated in bold,
    while the second best is underlined.}\label{tab:domainnet}
  \begin{tabular}{lllllllllllllll}
   \toprule
    \bfseries Method & \bfseries R$\to$C & \bfseries R$\to$P & \bfseries R$\to$S & \bfseries C$\to$R & \bfseries C$\to$P & \bfseries C$\to$S & \bfseries P$\to$R & \bfseries P$\to$C & \bfseries P$\to$S & \bfseries S$\to$R & \bfseries S$\to$C & \bfseries S$\to$P & \bfseries Avg \\
    \midrule
     ResNet-50 & 58.84 & 67.89 & 53.08 & 76.70 & 53.55 & 53.06 & 84.39 & 55.55 & 60.19 & 74.62 & 54.60 & 57.78 & 62.52 \\
     \hline
     UAN & 71.10 & 68.90 & 67.10 & 83.15 & 63.30 & 64.66 & 83.95 & 65.35 & 67.06 & 82.22 & 70.64 & 68.09 & 72.05 \\
     InstaPBM & 80.10 & 75.87 & 70.84 & 89.67 & 70.21 & 72.76 & 89.60 & 74.41 & 72.19 & 87.00 & 79.66 & 71.75 & 77.84 \\
     Sentry & \underline{83.89} & 76.72 & 74.43 & 90.61 & 76.02 & 79.47 & 90.27 & 82.91 & 75.60 & \underline{90.41} & 82.40 & 73.98 & 81.39 \\
     GSDE & 82.93 & \underline{79.16 }& \underline{80.76} & \underline{91.92} & \underline{78.16} & \underline{79.98} & \underline{90.92} & \underline{84.10} & \textbf{79.16} & 90.30 & 83.36 & \underline{76.07} & \underline{83.07} \\
     \hline
     A3 (Ours) & \textbf{94.85} & \textbf{92.15} & \textbf{86.48} & \textbf{95.40} & \textbf{94.74} & \textbf{88.46} & \textbf{94.05} & \textbf{96.1} & \underline{78.52} & \textbf{93.75} & \textbf{96.18} & \textbf{92.1} & \textbf{91.89} \\
    \bottomrule
  \end{tabular}
\end{table*}

\subsection{Ablation Studies}\label{sec:ablation}
We design different variants of the framework showcasing the
contributions of each component and the effect of each
active learning adaptation cycle on the performance of the target
model.

\paragraph{Varying acquisition strategies}
We evaluated the impact of a hybrid acquisition strategy on A3's
performance. Comparing uncertainty-only and random acquisition
functions, Table \ref{tab:ablation} demonstrates that the hybrid
strategy outperforms both. Notably, it significantly surpasses random
sampling and slightly improves upon the uncertainty-only function,
emphasizing the importance of combining diverse data samples with
informative ones.

\paragraph{Varying alignment techniques}
We examined the impact of individual domain alignment components on
A3. In Table \ref{tab:ablation}, we display the contributions of three
components, showing that the consolidated framework (Consolidated)
outperforms all variants. Notably, the variant with only domain
adversarial loss and virtual adversarial loss (DAL + VAT Only)
performs closely to the comprehensive framework compared to using only
entropy minimization loss (Entropy Only), emphasizing the influence of
adversarial losses on A3's performance.

\section{Limitations}
While A3 demonstrates strong performance across various domain adaptation tasks, it has limitations. Our method may struggle in scenarios with extreme domain shifts where low-level features differ significantly between source and target domains. For example, adapting between natural images and medical imagery could pose challenges.
Additionally, A3's effectiveness might be reduced when dealing with small target datasets, as the active learning component relies on a sufficiently large pool of unlabeled data to select informative samples.
Finally, like many deep learning approaches, A3 can be computationally intensive, especially during the iterative active learning cycles. This could limit its applicability in resource-constrained environments or real-time adaptation scenarios.

\section{Conclusion} \label{sec:conclusion}
We present A3, a novel framework for SFUDA that addresses two key
challenges: noisy pseudo-labels and distribution shift between source
and target domains. A3 utilizes self-supervised learning and active
adversarial training to tackle these issues. Specifically, we
introduce a domain adversarial classifier that aligns the marginal
feature distributions of the source and target domains and a virtual
adversarial loss which acts as a regularizer to prevent overfitting
and encourage model robustness. Furthermore, we propose an active
sampling strategy that computes the Shannon entropy of each target
sample to quantify the model's uncertainty. This uncertainty measure
is combined with k-means clustering to filter out only the most
informative and diverse samples for domain alignment. Through
extensive experiments, we demonstrate that A3 achieves superior
performance compared to existing UDA methods. In future work, it would
be interesting to explore applying A3's domain alignment and active
sampling techniques to other UDA approaches.

\section{Acknowledgment}
The results presented in this paper were obtained using
the Chameleon testbed supported by the National Science Foundation (NSF). Any opinions, findings, and conclusions or recommendations expressed in this material are those of the authors and do not necessarily reflect the views of the NSF.

\bibliographystyle{IEEEtran}
\bibliography{IEEEabrv,casl.bib}

\end{document}